\documentclass[10pt,twocolumn,letterpaper]{article}

\usepackage[pagenumbers]{cvpr} 

\definecolor{cvprblue}{rgb}{0.21,0.49,0.74}
\usepackage[pagebackref=true,breaklinks=true,colorlinks,bookmarks=false]{hyperref}

\usepackage{adjustbox}

\def\paperID{xxxx}
\def\confName{CVPR}
\def\confYear{2025}

\title{\textit{PassionSR}: Post-Training Quantization  with Adaptive Scale \\ in One-Step Diffusion based Image Super-Resolution}
\author{
\textbf{Libo Zhu}\textsuperscript{1},
\textbf{Jianze Li}\textsuperscript{1}, 
\textbf{Haotong Qin}\textsuperscript{2}\thanks{Corresponding authors: Haotong Qin and Yulun Zhang}, 
\textbf{Wenbo Li}\textsuperscript{3}, \\
\textbf{Yulun Zhang}\textsuperscript{1}\footnotemark[1],
\textbf{Yong Guo}\textsuperscript{4},
\textbf{Xiaokang Yang}\textsuperscript{1} \\
\textsuperscript{1}Shanghai Jiao Tong University, \textsuperscript{2}ETH Zürich,\\
\textsuperscript{3}Chinese University of Hong Kong,
\textsuperscript{4}Max Planck Institute for Informatics
}

\DeclareRobustCommand{\eg}{\emph{e.g.}}
\DeclareRobustCommand{\ie}{\emph{i.e.}}

\def\abstract{%
   \iftoggle{cvprpagenumbers}{}{%
     \thispagestyle{empty}
   }
   \centerline{\large\bf Abstract}%
   \vspace*{4pt}\noindent%
   \it\ignorespaces%
}

\begin{document}

\maketitle
\begin{abstract}
Diffusion-based image super-resolution (SR) models have shown superior performance at the cost of multiple denoising steps. However, even though the denoising step has been reduced to one, they require high computational costs and storage requirements, making it difficult for deployment on hardware devices. To address these issues, we propose a novel post-training quantization approach with adaptive scale in one-step diffusion (OSD) image SR, PassionSR. First, we simplify OSD model to two core components, UNet and Variational Autoencoder (VAE) by removing the CLIPEncoder. Secondly, we propose Learnable Boundary Quantizer (LBQ) and Learnable Equivalent Transformation (LET) to optimize the quantization process and manipulate activation distributions for better quantization. Finally, we design a Distributed Quantization Calibration (DQC) strategy that stabilizes the training of quantized parameters for rapid convergence. Comprehensive experiments demonstrate that PassionSR with 8-bit and 6-bit obtains comparable visual results with full-precision model. Moreover, PassionSR achieves significant advantages over recent leading low-bit quantization SR methods. Code will be at \url{https://github.com/libozhu03/PassionSR}.
\end{abstract}

\setlength{\abovedisplayskip}{2pt}
\setlength{\belowdisplayskip}{2pt}
\vspace{-7mm}
\section{Introduction}
\vspace{-2.5mm}
Image super-resolution (SR) is a fundamental and challenging task in computer vision, aiming to reconstruct high-resolution (HR) images from low-resolution (LR) inputs by recovering lost structures and details. Over time, various image SR models have been developed to tackle this challenge. Early image SR methods~\cite{ISR_using_conv,Zhang_2018_CVPR,chen2022CAT,chen2023DAT} focus on simple synthetic degradations, such as bicubic downsampling, to create LR-HR pairs. GAN approaches~\cite{BSRGAN,swinir,Real-ESRGAN} introduce more complex degradation processes. Although these methods have advanced image quality significantly, when applied to real-world datasets, they encounter some issues, such as training stability and performance drop, showing the limitations of early image SR methods.

\begin{figure}[t]
\scriptsize
\centering
\vspace{-3.5mm}
\begin{tabular}{c c c c c c c c}

\hspace{-0.44cm}

\begin{adjustbox}{valign=t}
\begin{tabular}{cccc}

\includegraphics[width=0.11\textwidth]{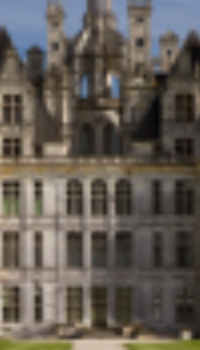} \hspace{-4.5mm} &
\includegraphics[width=0.11\textwidth]{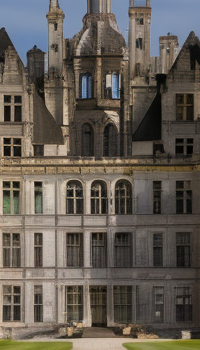} \hspace{-3.5mm} &
\includegraphics[width=0.11\textwidth]{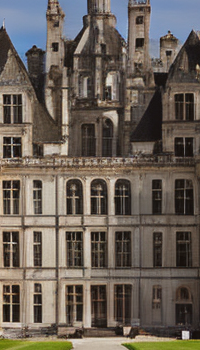} \hspace{-3.5mm} &
\includegraphics[width=0.11\textwidth]{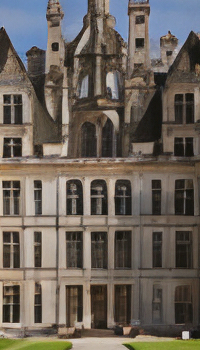} \hspace{-3.5mm} \\
LR ($\times$4) \hspace{-4.5mm} &
DiffBIR~\cite{diffbir} \hspace{-3.5mm} &
OSEDiff~\cite{OSEDiff} \hspace{-3.5mm} &
PassionSR (ours) \hspace{-3.5mm} \\
\# Step / Bits \hspace{-4.5mm} &
50 / 32-bit \hspace{-3.5mm} &
1 / 32-bit \hspace{-3.5mm} &
1 / 8-bit \hspace{-3.5mm} \\
Param. (M) / Ops (G) \hspace{-4.5mm} &
1,618 / 49,056 \hspace{-3.5mm} &
1,303 / 4,523 \hspace{-3.5mm} &
238 / 1,060 \hspace{-3.5mm} \\

\end{tabular}
\end{adjustbox}

\end{tabular}
\vspace{-3.5mm}
\caption{Visual comparison ($\times$4) between full-precision (FP) multi-step and one-step diffusion SR models and our 8-bit quantized PassionSR. Compared to FP models, PassionSR achieves about 81.77\% params reduction and 4$\times$ speedup.}
\label{fig:first-visual}
\vspace{-7.5mm}
\end{figure}

Recently, diffusion-based image SR models have been attracting researchers' attention. Diffusion models exhibit strong performance across various tasks, including image SR (see Fig.~\ref{fig:first-visual}), due to their excellent generation ability and robust training stability. Diffusion-based SR models~\cite{stablesr,diffbir,seesr} leverage their capability to capture complex data distributions, achieving superior perceptual quality.

However, achieving high-quality results with diffusion models comes at the expense of substantial computational demands, latency, and storage requirements. It poses significant challenges for deployment on hardware devices. Several strategies have been developed to improve the efficiency of diffusion models. For instance, fast diffusion samplers~\cite{song2022denoisingdiffusionimplicitmodels,zhao2023unipcunifiedpredictorcorrectorframework} and distillation techniques~\cite{song2023consistencymodels} have reduced the number of denoising steps. With the advance of score distillation-based methods~\cite{yin2024onestepdiffusiondistributionmatching,wang2023prolificdreamerhighfidelitydiversetextto3d}, one-step diffusion (OSD) image super-resolution (OSDSR) models have become feasible, such as SinSR~\cite{sinsr}, OSEDiff~\cite{OSEDiff}, and DFOSD~\cite{DFOSD}. Despite reducing denoising steps to one, OSDSR models still face high computational costs and storage requirements. For example, in OSEDiff~\cite{OSEDiff}, the parameters and Ops are 1,303M and 4,523G, respectively. The high computational complexity causes a significant burden for mobile devices like smartphones. This problem hinders the widespread use of diffusion models, especially in scenarios with limited storage and computational resources.

To compress the OSDSR model, quantization stands out as an effective approach. Model quantization~\cite{choukroun2019lowbitquantizationneuralnetworks,Ding_2022,pmlr-v139-hubara21a} is a powerful compression technique that reduces weights and activations from full-precision (FP) to low-bit precision. Thereby, it significantly lowers storing and computational demands by substituting floating-point operations with integer operations. However, performance gap between quantized and FP versions is inevitable. Minimizing this gap is crucial for the successful application of quantization techniques to OSDSR models.

\begin{table}[t]
\centering

\resizebox{1\columnwidth}{!}{
\begin{tabular}{c | c c c c | c}
\toprule
\rowcolor{cvprblue!30}
Components & UNet & VAE & DAPE& ClipEncoder & Total\\
\midrule
Params (M) & 865,785 & 83,614 & 160,335 & 193,055 & 1,302,789\\
\rowcolor{cvprblue!10}
MACs (G) &  339.241 & 1,781.123 & 126.591 & 14.856 & 2,261.811 \\
\bottomrule
\end{tabular}
}
\vspace{-3mm}
\caption{Params and FLOPs statistics in OSEDiff~\cite{OSEDiff}.}
\label{tab:OSEDiff params and FLOPs stastics}
\vspace{-7mm}
\end{table}

Although current low-bit quantization strategies~\cite{PTQ4DM,q-diffusion,q-dm,PTQD} have achieved promising results for multi-step diffusion quantization, significant performance drops still occur when applied to OSDSR models. Additionally, existing strategies are often hard to achieve optimal compression rates. We encounter three primary challenges: 

\emph{\textbf{\Rmnum{1}. Complex Model Structure.}} Unlike many multi-step diffusion models, the OSDSR model includes numerous submodules. \emph{\textbf{(\rmnum{1})}} In the previous multi-step diffusion quantization methods, most attention has been paid to UNet quantization while Variational Autoencoder (VAE) keeps FP. However, in OSDSR models, because UNet inference steps are reduced to one, it is essential to quantize VAE, which accounts for over 80\% of the computational load, as shown in Tab.~\ref{tab:OSEDiff params and FLOPs stastics}. \emph{\textbf{(\rmnum{2})}} We need to design special calibration strategies for branch modules (\eg, DAPE, CLIPEncoder), which brings lots of difficulties for quantization.

\begin{figure}[t]
\scriptsize
\centering
\begin{tabular}{c c c c c c c c}
\hspace{-0.44cm}

\begin{adjustbox}{valign=t}
\begin{tabular}{cccccc}

\includegraphics[width=0.11\textwidth]{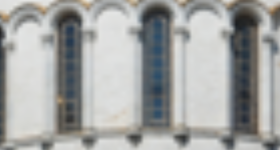} \hspace{-3.5mm} &
\includegraphics[width=0.11\textwidth]{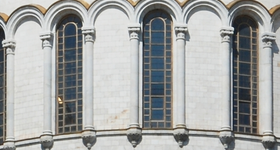} \hspace{-3.5mm} &
\includegraphics[width=0.11\textwidth]{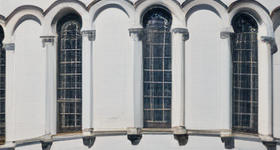} \hspace{-3.5mm} &
\includegraphics[width=0.11\textwidth]{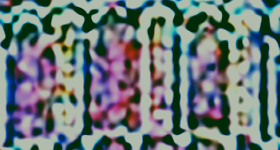} \hspace{-3.5mm} \\
LR ($\times$4) \hspace{-3.5mm} &
HR \hspace{-3.5mm} &
OSEDiff~\cite{OSEDiff} \hspace{-3.5mm} &
MaxMin~\cite{jacob2017quantizationtrainingneuralnetworks} \hspace{-3.5mm} \\
\includegraphics[width=0.11\textwidth]{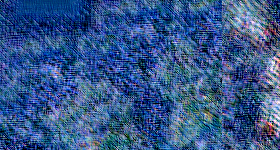} \hspace{-3.5mm} &
\includegraphics[width=0.11\textwidth]{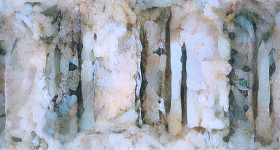} \hspace{-3.5mm} &
\includegraphics[width=0.11\textwidth]{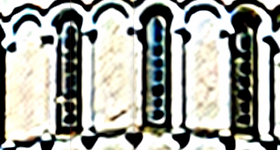} \hspace{-3.5mm} &
\includegraphics[width=0.11\textwidth]{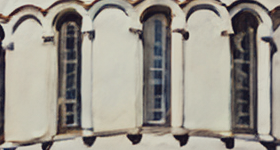} \hspace{-3.5mm} \\
LSQ~\cite{esser2019learned}  \hspace{-3.5mm} &
Q-Diffusion~\cite{q-diffusion} \hspace{-3.5mm} &
EfficientDM~\cite{EfficientDM}  \hspace{-3.5mm} &
PassionSR (ours) \hspace{-3.5mm}

\end{tabular}
\end{adjustbox}

\end{tabular}
\vspace{-3.5mm}
\caption{Visual comparison ($\times$4) of one-step diffusion SR models. We use OSEDiff as a 32-bit full-precision (FP) reference and provide 6-bit quantized version with different methods.}
\label{fig:visual-intro}
\vspace{-9.mm}
\end{figure}

\emph{\textbf{\Rmnum{2}. Transition from Multi-Step to One-Step.}} Most existing quantization techniques are designed for multi-step diffusion models and incorporate special techniques tailored to their multi-step nature. These techniques do not perform as effectively on OSDSR as on multi-step models. Many of them are even infeasible on OSDSR (see Fig.~\ref{fig:visual-intro}). It means we need to design a new calibration strategy tailored for OSDSR, taking better advantage of its features rather than applying the previous methods on OSDSR directly.

\emph{\textbf{\Rmnum{3}. Imbalanced Activation Distribution.}} Imbalanced activation distribution is a prevalent issue in model quantization. The presence of excessive outliers complicates the determination of optimal quantized parameters in traditional post-training quantization (PTQ) methods. And quantization-aware training (QAT) often encounters convergence challenges due to the quantization function and high demands in training time and memory usage.

Based on the above analyses, we propose a novel \emph{\textbf{P}}ost-training quantization method with \emph{\textbf{A}}daptive \emph{\textbf{S}}cale in one-\emph{\textbf{S}}tep diffus\emph{\textbf{ION}} based image \emph{\textbf{S}}uper-\emph{\textbf{R}}esolution, named \emph{\textbf{PassionSR}}. In this work, we select OSEDiff~\cite{OSEDiff} as our quantization backbone due to its excellent performance and high inference speed. \emph{\textbf{Firstly}}, we perform a pruning operation to simplify the model to its two core components, the UNet and VAE, with minimal or even no performance drop. We name the FP model structure after pruning as PassionSR-FP. It is easier for us to design calibration strategies. \emph{\textbf{Secondly}}, we propose a \emph{Learnable Boundary Quantizer} (LBQ) and \emph{Learnable Equivalent Transformation} (LET) to the quantization process. LBQ allows for training-based optimization. While, LET enables control over the distribution of activations without any additional computational expense, which is achieved through an adaptable scale parameter. The training process renders this scale parameter adaptive, which proves more effective than traditional initialization methods. \emph{\textbf{Thirdly}}, we design a calibration strategy, Distributed Quantization Calibration (DQC), that stabilizes the training of quantized parameters and promotes rapid convergence, achieving QAT-level effectiveness with PTQ efficiency.

\vspace{-2mm}
Comprehensive experiments (\ie, Figs.~\ref{fig:first-visual} and~\ref{fig:visual-intro}) indicate that PassionSR incurs a performance drop of less than 1\% at 8-bit precision and maintains relatively high performance even at 6-bit precision. Compared to recent leading diffusion quantization methods, PassionSR demonstrates significant performance advantages across various bit widths in one-step diffusion (OSD) model quantization. PassionSR delivers outstanding visual quality over other quantization methods. When compared with OSEDiff, our PassionSR achieves about 80$\sim$85\% parameters compression and operations reduction. Overall, our contributions are as follows:
\vspace{-2.5mm}
\begin{itemize}
    \item We propose a low-bit quantized OSDSR model, PassionSR. To the best of our knowledge, this is the first work to investigate low-bit quantization (\eg, 6-bit and 8-bit) for OSDSR in a PTQ manner.
    \item We design a UNet-VAE model (\ie, PassionSR-FP) structure that maintains high performance while simplifying the overall model to only UNet and VAE.
    \item We propose the Learnable Equivalent Transformation (LET) for the quantization of OSD models. Distributed Quantization Calibration (DQC) is designed to stabilize the training and accelerate the convergence.
    \item Our PassionSR achieves perceptual performance largely comparable to that of a full-precision model at 8-bit and 6-bit precision and obtains better performance and higher scores over other quantization methods.
\end{itemize}

\section{Related Work}
\vspace{-1mm}
\subsection{Single Image Super-Resolution}
\vspace{-1mm}
Single image super-resolution (SR) aims to recover high-resolution (HR) images from low-resolution (LR) inputs with unknown and complex degradation patterns. Numerous models have been developed to address this challenge. In addition to early SR models~\cite{AISR,zhang2018imagesuperresolutionusingdeep,chen2023crossaggregationtransformerimage} and GAN-based approaches~\cite{BSRGAN,Real-ESRGAN,swinir}, stable diffusion (SD)~\cite{rombach2022highresolutionimagesynthesislatent} has emerged as a powerful technique due to its robust capability in capturing complex data distributions and providing strong generative priors. Related methods, including StableSR~\cite{stablesr}, DiffBIR~\cite{diffbir}, and SeeSR~\cite{seesr}, enhance the perceptual quality of generated images. However, their multi-step processes introduce higher latency, which hinders real-time applications. To address this limitation, one-step diffusion (OSD) models, such as SinSR~\cite{sinsr} and OSEDiff~\cite{OSEDiff}, have been developed to reduce inference latency by accelerating the process to a single step.
\vspace{-1mm}
\subsection{Model Quantization}
\vspace{-1mm}
Model quantization is a critical technique for accelerating models by reducing computational costs and inference time. Depending on whether the model’s weights are retrained, quantization methods are divided into two categories: post-training quantization (PTQ) and quantization-aware training (QAT). PTQ is highly time-efficient as it only calibrates the quantized parameters rather than finetunes the entire model. ZeroQuant~\cite{ZeroQuant} calibrates quantized parameters without additional calibration datasets, and BRECQ~\cite{brecq} introduces a block-wise reconstruction PTQ method. QAT can achieve higher accuracy but incurs high training costs. As a representative quantization method, LSQ~\cite{esser2019learned} improves low-bit quantization with a learnable step size. 

\vspace{-1mm}
\subsection{Quantization of Diffusion Models}
\begin{figure}[t]
\centering
\includegraphics[width=1\columnwidth]{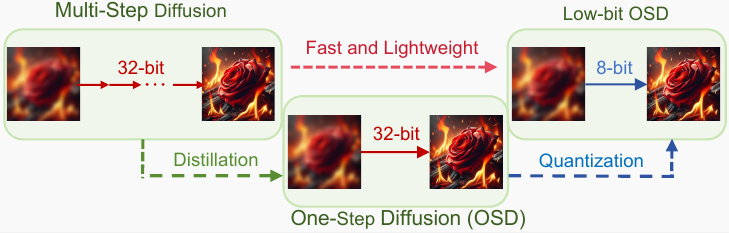}
\vspace{-7mm}
\caption{Diffusion-based image SR acceleration.}
\label{fig:progress}
\vspace{-6.5mm}
\end{figure}

\vspace{-1mm}
As diffusion models evolve rapidly, researchers have focused on improving their efficiency through quantization. PTQ4DM~\cite{PTQ4DM} first investigates quantized diffusion models, identifying key challenges to overcome. Further works, including Q-Diffusion~\cite{q-diffusion}, PTQD~\cite{PTQD}, and QAT methods like Q-DM~\cite{q-dm}, have made significant progress by developing specialized calibration strategies tailored to diffusion models. Notably, TDQ~\cite{tdq} utilizes an MLP layer to predict quantized parameters, and APQ-DM~\cite{APQ-DM} designs a distribution-aware quantization approach to minimize quantization error. Additionally, QALoRA~\cite{qalora}, is a notable quantization method for large language models, reducing quantization error by finetuning LoRA layers along with quantized parameters. It is adopted to quantize diffusion models in EfficientDM~\cite{EfficientDM}. QuEST~\cite{QuEST} finds that layers like the feedforward layer are sensitive to quantization. QuEST improves performance by selectively retraining these layers. These methods have advanced low-bit quantization for multi-step diffusion models. However, there are few works specifically addressing the low-bit quantization of one-step diffusion (OSD) models, which are significantly different from multi-step diffusion models. We follow the strategy in Fig.~\ref{fig:progress} to accelerate the diffusion-based SR models, especially the OSD models.

\begin{figure*}[t]
\centering
\includegraphics[width=\textwidth]{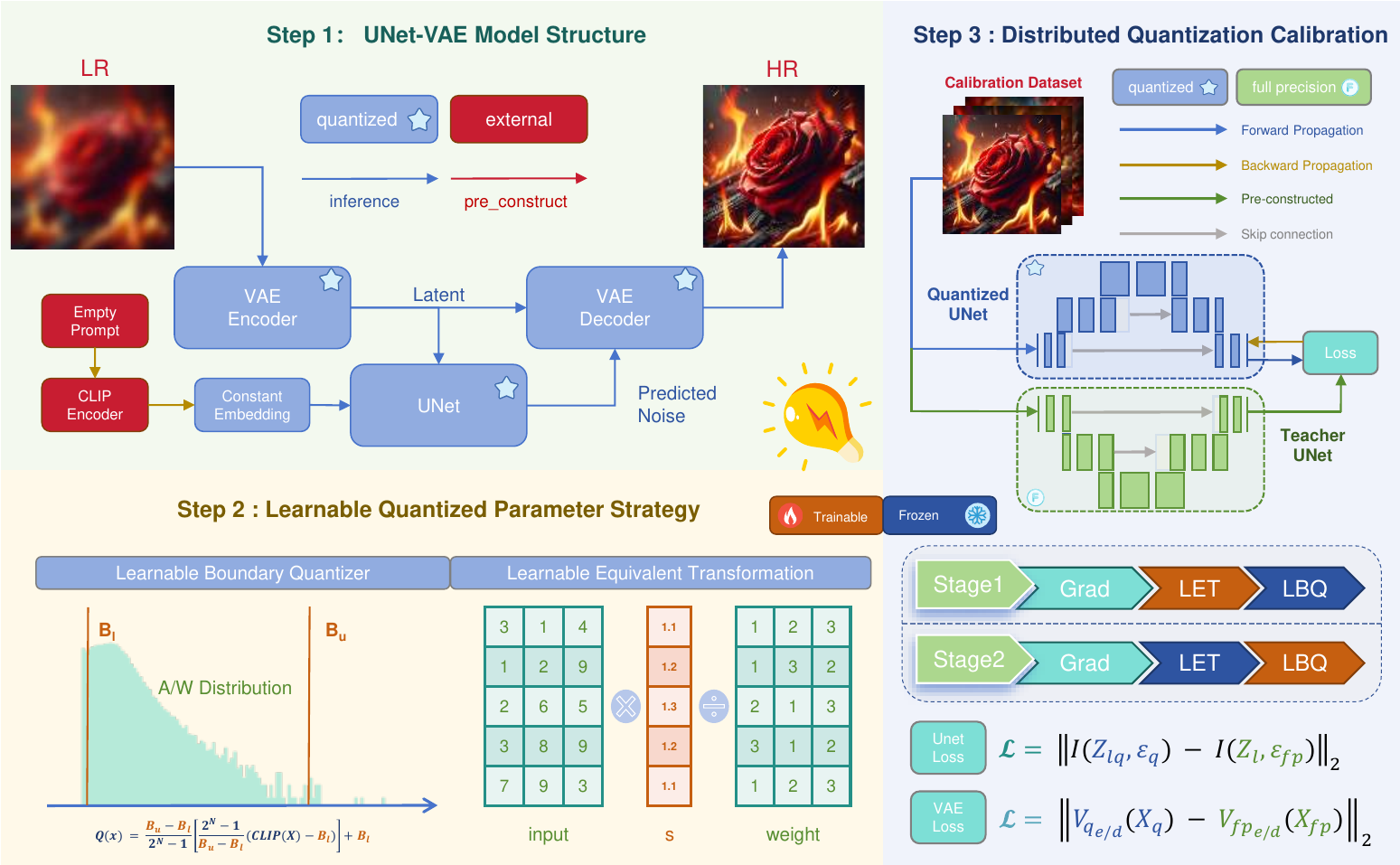}
\vspace{-6mm}
\caption{Overview of our PassionSR. {Step 1}: we simplify OSEDiff~\cite{OSEDiff} by removing DAPE and CLIP Encoder, obtaining PassionSR-FP. {Step 2}: the quantizer we use has two key trainable parts, consisting of the Learnable Boundary Quantizer and Learnable Equivalent Transformation. {Step 3}: we design a distributed calibration strategy and special loss function to accelerate convergence of calibration.}
\label{fig:overview}
\vspace{-5mm}
\end{figure*}

\vspace{-4mm}
\section{Methods}
\vspace{-1.5mm}
\subsection{Preliminaries}
\vspace{-1.5mm}
\noindent \textbf{Diffusion Models.} Diffusion models~\cite{rombach2022highresolutionimagesynthesislatent} are generative techniques that gradually introduce noise into data and then learn to invert this process to create new samples. The process starts with a real data distribution $ p_{\text{data}}(\mathbf{x}) $, where data $ \mathbf{x}_0 $ is gradually transformed into noise over several time steps, $ t = 1, 2, \dots, T $. At each step, the data evolves towards randomness with noise controlled by a time-dependent parameter $ \beta_t $, which increases with each step. After enough steps, the data distribution becomes close to the standard normal distribution, i.e., $ \mathbf{x}_t \sim \mathcal{N}(0, \mathbf{I}) $. Then process is reversed to recover the original data $ \mathbf{x}_0 $ from the noisy data $ \mathbf{x}_t $ by training a neural model to predict $ \mathbf{x}_{t-1} $ from $ \mathbf{x}_t $. The network learns the denoising function, parameterized by $ \theta $, which predicts the clean data mean $ \mu_\theta(\mathbf{x}_t, t) $ and the noise level $ \sigma^2(t) $. By iterating this process, new samples are generated from random noise, gradually refined step by step. This denoising process allows the model to generate highly realistic samples.

\noindent \textbf{Model Quantization.} Model quantization uses both the scale factor and zero point bias to handle the shift in the data distribution. It reduces memory consumption and computation time by mapping model parameters and activations to low-bit integers. Given a floating-point vector $ \mathbf{x} $, the quantization operation is as follows:
\begin{equation}
    \hat{x} = \text{Q}(\mathbf{x}, s, z) = s \cdot \text{Clip}\left(\frac{\mathbf{x}-z}{s}, l, u \right)+z,
\end{equation}
where $ s $ is the scaling factor that controls quantization precision, and the zero-point bias $ z $ shifts the data before scaling. $ \text{Clip}(\cdot, l, u) $ bounds the quantized values within the range from the lower bounds $l$ to upper bounds $u$.

Since quantization involves the non-differentiable rounding operations, the straight-through estimator (STE~\cite{liu2022nonuniform}) is commonly used to approximate gradients:
\begin{equation}
\frac{\partial L(x)}{\partial x} \approx 
\begin{cases} 
1 & \text{if } x \in [l, u], \\
0 & \text{otherwise}.
\end{cases}
\label{equ:STE}
\end{equation}

\subsection{UNet-VAE Model Structure}
\vspace{-1mm}
Based on OSEDiff~\cite{OSEDiff}, we obtain a full-precision (FP) model PassionSR-FP by simplifying the original design to only UNet and VAE while maintaining comparable performance. The details of the simplified model structure are shown in Step 1 of Fig.~\ref{fig:overview}. Compared to OSEDiff, we replace the DAPE-CLIPEncoder branch with a constant embedding, preprocessed by the ClipEncoder using an empty string. Owing to the similar basic components, we can adopt similar calibration strategy on them.

\vspace{-2.5mm}
\subsection{Learnable Quantized Parameter Strategy}
\vspace{-2mm}
To minimize performance drop in quantization, we introduce learnable quantized parameters and use a pre-constructed small calibration dataset to guide the training, which is both time- and memory-efficient. It is important to note that only the quantized parameters are trained, while the original weight parameters remain unchanged. Step 2 in Fig.~\ref{fig:overview} visualizes the two key components with trainable parameters of our quantizer, LBQ and LET.

\vspace{-5mm}
\subsubsection{Learnable Boundary Quantizer (LBQ)} 
\label{sec: Learnable Boundary Quantizer}
\vspace{-2mm}
To simulate the quantization loss, we apply fake quantization~\cite{jacob2017quantizationtrainingneuralnetworks} to both activations and weights. The quantization and dequantization processes are defined as Eq.~\eqref{equ:fake_quantization}. Using it we define the Learnable Boundary Quantizer
\begin{equation}
\begin{cases}
    X_{\text{c}} = \textbf{Clip}(X, B_{\text{l}}, B_{\text{u}}),  \alpha = \dfrac{B_{\text{u}} - B_{\text{l}}}{2^N - 1}, 
    \beta = B_{\text{l}}, \\
    X_{\text{I}} = \left\lfloor \dfrac{X_{\text{c}} - \beta}{\alpha} \right\rceil, X_{\text{q}} = \alpha X_{\text{I}} + \beta,
\end{cases}
\label{equ:fake_quantization}
\end{equation}
where $X_{\text{q}}$ is the fake-quantized value used to simulate quantization error. The function $\textbf{Clip}(X, B_{\text{l}}, B_{\text{u}})$ is defined as $\max(\min(X, B_{\text{u}}), B_{\text{l}})$, and $\left\lfloor x \right\rceil$ rounds the input $x$ to the nearest integer. The parameters $B_{\text{l}}$ and $B_{\text{u}}$, representing the lower and upper boundaries, are the only trainable parameters, deciding the function of whole LBQ.

\vspace{-5mm}
\subsubsection{Learnable Equivalent Transformation (LET)}
\vspace{-2mm}
Inspired by SmoothQuant~\cite{smoothquant} and OmniQuant~\cite{OmniQuant}, we introduce channel-wise trainable scale and shift factors to adjust the activation distribution, effectively addressing challenges posed by outliers during quantization.

The fundamental layers to be quantized in diffusion models include the linear layer, convolution layer, and matrix multiplication within the attention layer. We apply equivalent transformations to these layers, balancing the distribution of activations and weights and making the model more suitable for quantization. To maximize its potential to reduce performance drop, finetuning is required to obtain better parameters in LET. 

\noindent\textbf{Linear Layer.} In a typical linear layer, whose input dimension is $C_\text{in}$ and output dimension is $C_\text{out}$, the input matrix $X \in \mathbb{R}^{N \times C_\text{in}}$ is multiplied by the weight matrix $W \in \mathbb{R}^{C_\text{in} \times C_\text{out}}$, along with a bias matrix $B \in \mathbb{R}^{1 \times C_\text{out}}$, resulting in an output matrix $Y \in \mathbb{R}^{N \times C_\text{out}}$.

To introduce equivalent transformations in the linear layer, we apply a learnable scale factor $s \in \mathbb{R}^{1 \times C_\text{in}}$ and an offset factor $\delta \in \mathbb{R}^{1 \times C_\text{in}}$ to the input $X$. To preserve the output $Y$, corresponding transformations are applied to $W$ and $B$, resulting in transformed input $\tilde{X}$, weight $\tilde{W}$, and bias $\tilde{B}$, as shown in Eq.~\eqref{linear transformation}:
\begin{equation}
\tilde{W} = s \odot W, \tilde{X} = (X - \delta) \oslash s, \tilde{B} = B + \delta W,
\label{linear transformation}
\end{equation}
where $\odot$, $\oslash$ are element-wise multiplication and division.

\vspace{-0.8mm}
After applying the quantizer LBQ to the linear layer, the simulated quantized output $Y_{\text{q}}$ is expressed as Eq.~\eqref{equ:linear_quantization}. And Eq.~\eqref{equ:fp_linear} refers to full precision output $Y_{\text{fp}}$:
\begin{numcases}{\text{}} 
Y_{\text{q}} = Q_{\text{a}}(\tilde{X}) Q_{\text{w}}(\tilde{W}) + Q_{\text{w}}(\tilde{B}) \label{equ:linear_quantization},\\
Y_{\text{fp}} = \tilde{X}\tilde{W} + \tilde{B} = XW + B = Y, \label{equ:fp_linear} 
\end{numcases}
where $ Q_{\text{a}}$ and $ Q_{\text{w}}$ are the activation and weight quantizers.

\noindent\textbf{Convolution Layer.} Equivalent transformations are applied along the channel dimension. The operations and quantization methods are similar to those in the linear layer by replacing matrix multiplication with convolution.

\noindent\textbf{Attention Operation.} In diffusion models, the attention operation is crucial for quantization due to the emergence of outliers following the softmax operation. It has been observed in vision transformers~\cite{ptq4vit}. We apply similar transformations to the matrix multiplication of the $Q$, $K$, and $V$ matrices in the transformer blocks. For example, the multiplication of the $Q$ and $K$ matrices can be transformed as shown in Eq.~\eqref{equ:QK_trans}. After this transformation, the quantizer is applied to the transformed matrices $\tilde{Q}$ and $\tilde{K}$, yielding the simulated quantized output $P_{\text{q}}$:
\begin{equation}
\begin{cases}
    \tilde{Q} = Q \oslash s, \quad s \odot K = \tilde{K}, \\
    P_{\text{q}} = \text{Softmax}(Q_{\text{a1}}(\tilde{Q}) Q_{\text{a2}}(\tilde{K}^\top)).
\end{cases}
\label{equ:QK_trans}
\end{equation}
It is worth noting that the scale factor $s$ and offset $\delta$ used in activation conversion can be merged into the preceding linear, convolution, or normalization layers. $s$ and $\delta$ used in weight and bias conversion can be incorporated directly into the weights and biases. This integration brings no additional memory or computation costs, making it hardware-friendly. Hardware experiments by AWQ~\cite{AWQ} have confirmed the compatibility of this equivalent transformation when employing quantized models to hardware devices. 

\vspace{-2mm}
\subsection{Quantization Calibration Design}
\vspace{-1mm}
Obtaining optimal quantized parameters is critical in any quantization task. With trainable parameters in our quantizers, we design a quantization calibration strategy to minimize quantization error with low latency and memory costs. The calibration pipeline is shown in Step 3 of Fig.~\ref{fig:overview}.

\vspace{-4mm}
\subsubsection{Distributed Quantization Calibration (DQC)}
\begin{figure}[t]
\centering
\vspace{1.5mm}
\includegraphics[width=0.98\columnwidth, height=0.5\columnwidth]{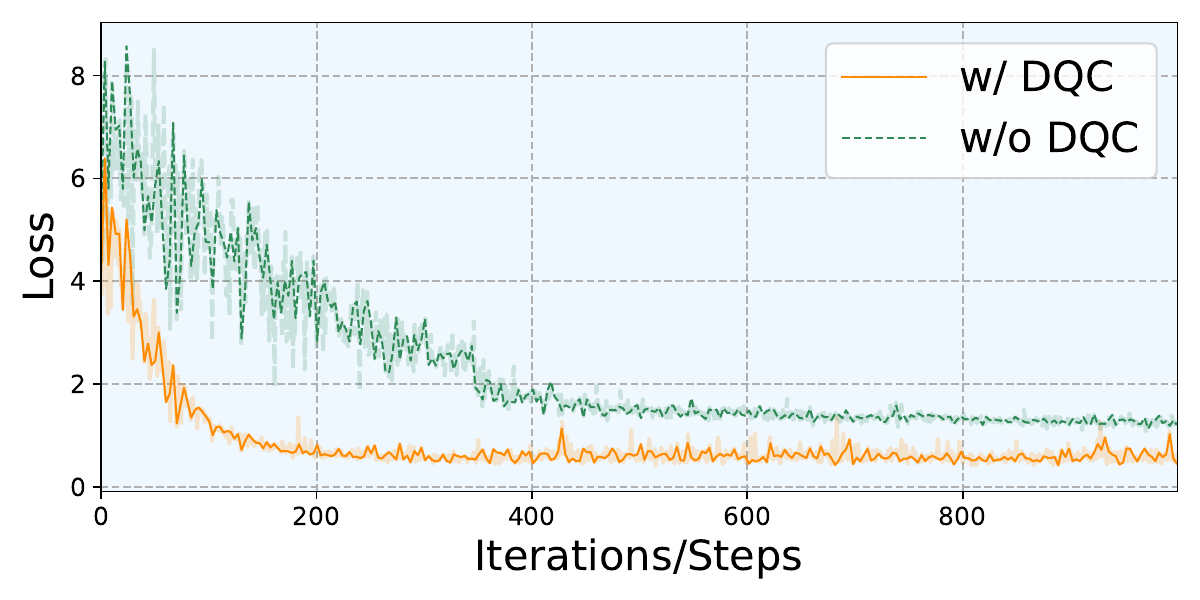}
\vspace{-4mm}
\caption{Loss comparison between w/ and w/o DQC}
\label{fig:loss_comparison}
\vspace{-7mm}
\end{figure}

\vspace{-3mm}
Given the properties of the rounding function, the training process in model quantization tends to be unstable. This instability will get worse when calibrating the boundaries in LBQ and the scale factor in LET at the same time. Therefore, we propose a Distributed Quantization Calibration (DQC) strategy that divides the entire calibration process into two stages (\ie, Step 3 of Fig.~\ref{fig:overview}). We reinitialize the LBQ to adapt to the new vector for quantization when the scale factors and offsets in LET are updated in Stage 1. Our DQC strategy significantly accelerates convergence and stabilizes the training process, as shown in Fig.~\ref{fig:loss_comparison}. 

\vspace{-7mm}
\subsubsection{Loss Design for UNet and VAE}
\vspace{-3mm}
We further use a model-wise quantization calibration strategy, which ensures that each module’s quantized output serves as input for the next module. It can avoid the accumulation of quantization errors across different modules.

The design of the loss function is crucial for calibrating quantized parameters. We use different loss functions for the UNet and VAE. We design the loss functions for the VAE encoder and decoder expressed as follows:
\begin{equation}
    \begin{cases}
        \mathcal{L}_{\text{VAE}_{\text{e}}} &= \Vert V_{\text{q}_{\text{e}}}(X_{\text{fp}}) - V_{\text{fp}_{\text{e}}}(X_{\text{fp}}) \Vert_2, \\
        \mathcal{L}_{\text{VAE}_{\text{d}}} &= \Vert V_{\text{q}_{\text{d}}}(X_{\text{q}}) - V_{\text{fp}_{\text{d}}}(X_{\text{fp}}) \Vert_2,
    \end{cases}
\label{loss_VAE}
\end{equation}
where $V_{\text{q}}$ denotes the quantized VAE, $V_{\text{fp}}$ represents the full-precision VAE, and $\text{[]}_{\text{e}}/\text{[]}_{\text{d}}$ indicates the encoder or decoder of the VAE. $\Vert \cdot \Vert_2$ represents the mean square error (MSE) loss. $X_{\text{q}}$ is the quantized input, the output of the previous quantized modules, while $X_{\text{fp}}$ is the full-precision input, the output of the previous full-precision modules.

For one-step diffusion (OSD) models, the time-step and noise level ($1 - \hat{\alpha}$) are constant. So the transformation function $I(Z_{\text{l}}, \varepsilon)$ from the predicted noise $\varepsilon(Z_{\text{l}})$ and input latent feature $Z_{\text{l}}$ to the output latent feature $Z_h$ can be defined as:
\begin{equation}
    I(Z_{\text{l}}, \varepsilon) = Z_{\text{h}} = \sqrt{\frac{1}{\hat{\alpha}}}Z_{\text{l}} - \sqrt{\frac{1 - \hat{\alpha}}{\hat{\alpha}}} \varepsilon(Z_{\text{l}}).
\label{get_x0}
\end{equation}

We use MSE in latent feature space, leveraging the transformation function $I(Z_\text{l}, \varepsilon)$ to facilitate smoother gradient descent and faster model convergence. We design the loss function for UNet as follows: 
\begin{equation}
    \mathcal{L}_\text{Unet} = \Vert I(Z_{\text{lq}}, \varepsilon_{\text{q}}) - I(Z_{\text{l}}, \varepsilon_{\text{fp}}) \Vert_2.
\label{loss_unet}
\end{equation}

\begin{table*}[htbp]
\centering
\scriptsize
\resizebox{\textwidth}{!}{
\begin{tabular}{c | c | c | c c c c c c c c}
\toprule
\rowcolor{cvprblue!30}
Datasets & Bits & Methods & PSNR$\uparrow$  & SSIM$\uparrow$
& LPIPS$\downarrow$ & DISTS$\downarrow$ &  NIQE$\downarrow$
& MUSIQ$\uparrow$ & MANIQA$\uparrow$ & CLIP-IQA$\uparrow$\\
\midrule
& \multirow{2}{*}{W32A32} 
&  OSEDiff~\cite{OSEDiff} &25.27 & 0.7379 & 0.3027	& 0.1808 &4.355	& 67.43	&0.4766	&0.6835  \\
&  & PassionSR-FP & 25.39 & 0.7460 & 0.2984 & 0.1813 & 4.453 & 67.05 & 0.4680 & 0.6796 \\
\cline{2-11}
& \multirow{5}{*}{W8A8} 
& MaxMin~\cite{jacob2017quantizationtrainingneuralnetworks} & 23.16	&0.6875	&0.5463	&0.2879	&7.932	&32.92	&0.1849	&0.2363\\
& &LSQ~\cite{esser2019learned} & 15.39 & 0.3375 & 0.9944 & 0.5427 & 10.08 & 50.11 & 0.3533 & 0.3173 \\
& & Q-Diffusion~\cite{q-diffusion} & 24.88 & 0.6967 & 0.4993 & 0.2696 & 8.437 & 44.69 & 0.2352 & 0.5604 \\
& & EfficientDM~\cite{EfficientDM} &14.77	&0.4253	&0.5478	&0.3462	&7.526	&44.75	&0.2568	&0.4000 \\
\rowcolor{cvprblue!10}
\cellcolor{white}& \cellcolor{white} & PassionSR (ours) &\textcolor{red}{25.67}	&\textcolor{red}{0.7499}	&\textcolor{red}{0.3140}	&\textcolor{red}{0.1932}	&\textcolor{red}{5.654}	&\textcolor{red}{65.88}	& \textcolor{red}{0.4437}	&\textcolor{red}{0.6912}\\
\cline{2-11}
& \multirow{5}{*}{W6A6} 
& MaxMin~\cite{jacob2017quantizationtrainingneuralnetworks} &15.55	&0.2417	&0.8018	&0.4449	&9.263	&42.15	&0.2791	&0.4174\\
& &LSQ~\cite{esser2019learned} & 13.73 & 0.1081 & 1.0900 & 0.5450 & 8.430 & 53.61 & 0.3036 & 0.4396 \\
& & Q-Diffusion~\cite{q-diffusion} & 19.75 & 0.4727 & 0.6877 & 0.4024 &\textcolor{red}{7.381} & \textcolor{red}{56.46} & \textcolor{red}{0.4380} & \textcolor{red}{0.6439} \\
& & EfficientDM~\cite{EfficientDM} &14.75	&0.4386	&0.5233	&0.3451	&7.497	&42.97	&0.2498	&0.3740 \\
\rowcolor{cvprblue!10}
\cellcolor{white}\multirow{-12}{*}{RealSR} & \cellcolor{white} & PassionSR (ours) & \textcolor{red}{25.15}	& \textcolor{red}{0.7196}	& \textcolor{red}{0.4199}	&\textcolor{red}{0.2592}	&8.618	&44.43	&0.2131	&0.4612 \\
\midrule
& \multirow{2}{*}{W32A32} 
& OSEDiff~\cite{OSEDiff} & 25.57    & 0.7885    & 0.3447 & 0.1808 & 4.371 & 37.22 & 0.4794 & 0.7540 \\
& & PassionSR-FP & 26.70	&0.7978	&0.3339	&0.1765	&4.336	&37.03	&0.4686	&0.7520 \\
\cline{2-11}
& \multirow{5}{*}{W8A8} 
& MaxMin~\cite{jacob2017quantizationtrainingneuralnetworks}&24.97	&0.7989	&0.5091	&0.2921	&8.215	&24.05	&0.1846	&0.3163\\
& &LSQ~\cite{esser2019learned} & 14.56 & 0.1795 & 1.1661 & 0.592 & 10.19 & 29.07 & 0.4010 & 0.3970 \\
& & Q-Diffusion~\cite{q-diffusion} & 27.14 & 0.7184 & 0.4765 & 0.2895 & 9.861 & 26.44 & 0.2284 & 0.5608 \\
& & EfficientDM~\cite{EfficientDM} &15.55	&0.4183	&0.6291	&0.3555	&6.859	&28.61	&0.2468	&0.4150\\
\rowcolor{cvprblue!10}
\cellcolor{white}& \cellcolor{white} & PassionSR (ours) & \textcolor{red}{27.41}	&\textcolor{red}{0.8146}	&\textcolor{red}{0.3422}	&\textcolor{red}{0.1918}	&\textcolor{red}{6.070}	&\textcolor{red}{33.56}	&\textcolor{red}{0.4286}	&\textcolor{red}{0.7554} \\
\cline{2-11}
& \multirow{5}{*}{W6A6} 
& MaxMin~\cite{jacob2017quantizationtrainingneuralnetworks}& 13.08	&0.2291	&0.8131	&0.5077	&10.51	& \textcolor{red}{35.83}	&0.2702	&0.3864 \\
& &LSQ~\cite{esser2019learned} & 12.95 & 0.0934 & 1.1890 & 0.5833 & 8.591 &26.39 & 0.2911 & 0.5600 \\
& & Q-Diffusion~\cite{q-diffusion} & 21.75 & 0.6096 & 0.7008 & 0.4039 & 6.854 & 24.39 & \textcolor{red}{0.4109} & \textcolor{red}{0.6696} \\
& & EfficientDM~\cite{EfficientDM} &15.07	&0.4287	&0.6127	&0.357	& \textcolor{red}{6.690}	&28.37	&0.2351	&0.3973 \\
\rowcolor{cvprblue!10}
\cellcolor{white} \multirow{-12}{*}{DRealSR} & \cellcolor{white} & PassionSR (ours) & \textcolor{red}{26.62}	&\textcolor{red}{0.7984}	&\textcolor{red}{0.4429}	&\textcolor{red}{0.2571}	&8.484	&26.26	&0.1824	&0.4358 \\
\midrule
& \multirow{2}{*}{W32A32} 
& OSEDiff~\cite{OSEDiff} & 24.95 & 0.7154	& 0.2325	& 0.1197	& 3.616 &	68.92	& 0.4340 &	0.6842 \\
& & PassionSR-FP & 25.16	&0.7221	&0.2373	&0.1185	&3.573	&69.27	&0.4402	&0.6958\\
\cline{2-11}
& \multirow{5}{*}{W8A8} 
& MaxMin~\cite{jacob2017quantizationtrainingneuralnetworks} & 22.33	&0.6618	&0.5639	&0.2731	&7.563	&33.68	&0.1913	&0.2818\\
& &LSQ~\cite{esser2019learned} & 13.90 & 0.2537 & 0.9932 & 0.5515 & 9.578 & 48.11 & 0.3512 & 0.3246 \\
& &  Q-Diffusion~\cite{q-diffusion} & 24.20 & 0.6813 & 0.3997 & 0.2400 & 7.955 & 51.95 & 0.2709 & 0.6243 \\
& & EfficientDM~\cite{EfficientDM}& 15.24	&0.4954	&0.6041	&0.3374	&6.856	&48.78	&0.2685	&0.4235 \\
\rowcolor{cvprblue!10}
\cellcolor{white}& \cellcolor{white} & PassionSR (ours) & \textcolor{red}{25.11}	&\textcolor{red}{0.7199}	&\textcolor{red}{0.2496}	&\textcolor{red}{0.1277}	&\textcolor{red}{4.424}	&\textcolor{red}{67.92}	&\textcolor{red}{0.3993}	&\textcolor{red}{0.6939}\\
\cline{2-11}
& \multirow{5}{*}{W6A6}
& MaxMin~\cite{jacob2017quantizationtrainingneuralnetworks} &11.66	&0.1606	&0.8509	&0.4966	&11.30	&45.47	&0.2764	&0.3523\\
& &LSQ~\cite{esser2019learned} & 12.21 & 0.0858 & 1.0695 & 0.5424 & 8.564 &\textcolor{red}{52.74} & 0.2872 & 0.4692 \\
& &  Q-Diffusion~\cite{q-diffusion} & 18.92 & 0.4939 & 0.6227 & 0.3718 & \textcolor{red}{6.162}& 51.50 & \textcolor{red}{0.3946} & \textcolor{red}{0.5814} \\
& & EfficientDM~\cite{EfficientDM} &15.09	&0.4991	&0.5953	&0.3292	&6.900	&46.01	&0.2570	&0.4007 \\
\rowcolor{cvprblue!10}
\cellcolor{white}\multirow{-12}{*}{DIV2K\_val} & \cellcolor{white} & PassionSR (ours) & \textcolor{red}{24.34}	& \textcolor{red}{0.7097}	& \textcolor{red}{0.3440}	& \textcolor{red}{0.2075}	& 7.039	&51.19	&0.2267	&0.4802 \\
\bottomrule
\end{tabular}
}
\vspace{-3mm}
\caption{Quantitative UNet-VAE quantization experiments results. PassionSR-FP is used as full-precision backbones rather than original OSEDiff. W8A8 denotes 8 bit weight and 8 bits activation quantization. The best results in the same setting are colored with \textcolor{red}{red}.}
\label{tab:Whole model quantization experiment results.}
\vspace{-6mm}
\end{table*}

\begin{figure*}[!t]
\scriptsize
\centering
\begin{tabular}{cccccccc}
\hspace{-0.44cm}
\begin{adjustbox}{valign=t}
\begin{tabular}{c}
\includegraphics[width=0.185\textwidth]{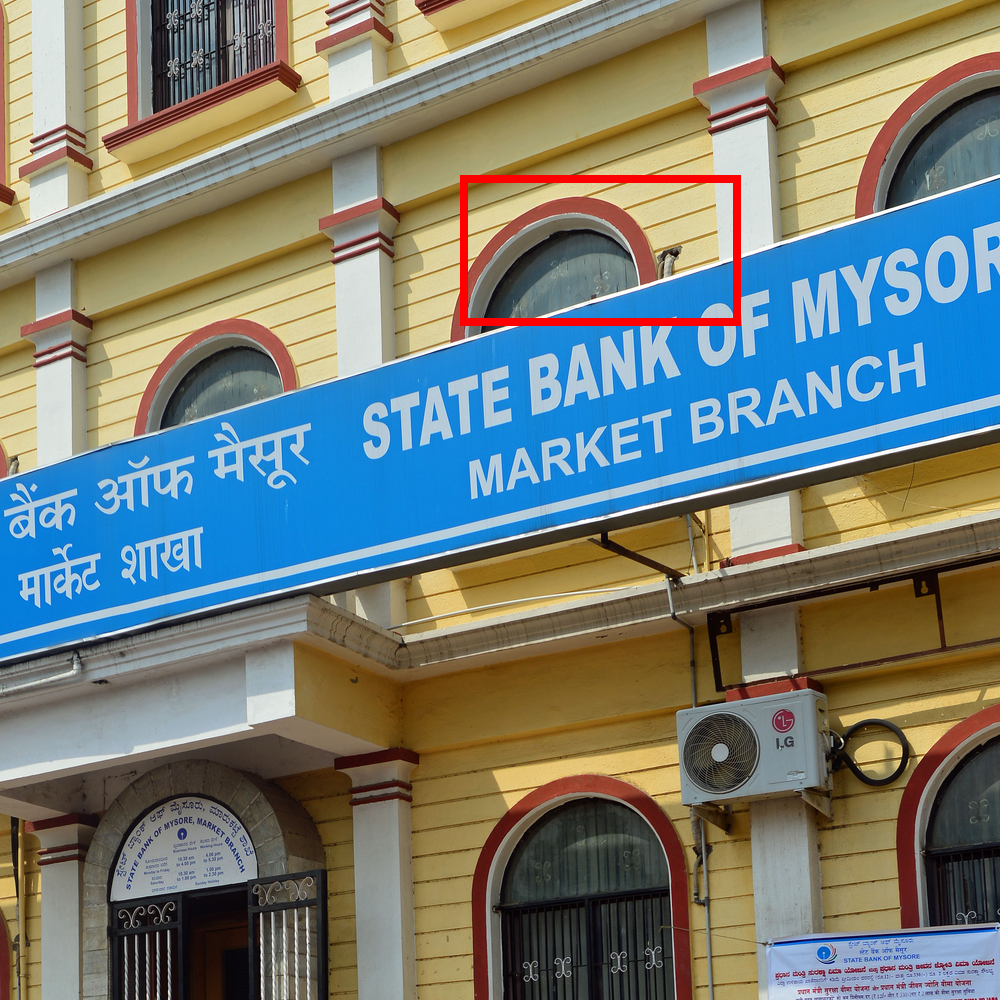}
\\
DIV2K\_val: 0891
\end{tabular}
\end{adjustbox}
\hspace{-0.46cm}
\begin{adjustbox}{valign=t}
\begin{tabular}{cccccc}
\includegraphics[width=0.154\textwidth]{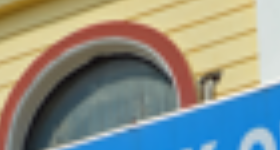} \hspace{-3.5mm} &
\includegraphics[width=0.154\textwidth]{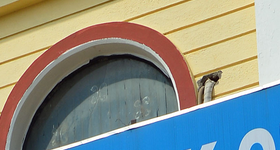} \hspace{-3.5mm} &
\includegraphics[width=0.154\textwidth]{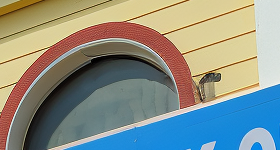} \hspace{-3.5mm} &
\includegraphics[width=0.154\textwidth]{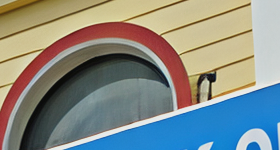} \hspace{-3.5mm} &
\includegraphics[width=0.154\textwidth]{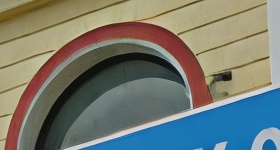} \hspace{-3.5mm} &
\\ 
LR \hspace{-3.5mm} &
HR \hspace{-3.5mm} &
DiffBIR~\cite{diffbir} / 32-bit \hspace{-3.5mm} &
OSEDiff~\cite{OSEDiff}  / 32-bit\hspace{-3.5mm} &
PassionSR-FP / 32-bit \hspace{-3.5mm} &
\\
\includegraphics[width=0.154\textwidth]{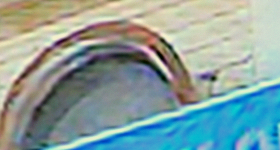} \hspace{-3.5mm} &
\includegraphics[width=0.154\textwidth]{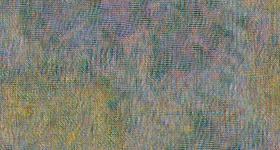} \hspace{-3.5mm} &
\includegraphics[width=0.154\textwidth]{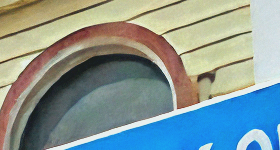} \hspace{-3.5mm} &
\includegraphics[width=0.154\textwidth]{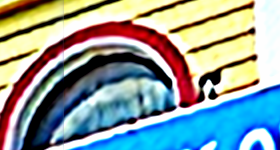} \hspace{-3.5mm} &
\includegraphics[width=0.154\textwidth]{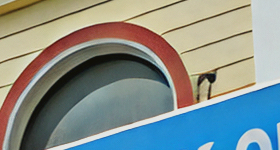} \hspace{-3.5mm} &
\\ 
MaxMin~\cite{choukroun2019lowbitquantizationneuralnetworks} / 8-bit \hspace{-3.5mm} &
LSQ~\cite{esser2019learned}  / 8-bit \hspace{-3.5mm} &
Q-Diffusion~\cite{q-diffusion}  / 8-bit \hspace{-3.5mm} &
EfficientDM~\cite{EfficientDM}  / 8-bit \hspace{-3.5mm} &
PassionSR (ours)  / 8-bit \hspace{-3.5mm}
\\
\end{tabular}
\end{adjustbox}
\end{tabular}
\vspace{-0.5mm}

\begin{tabular}{cccccccc}
\hspace{-0.44cm}
\begin{adjustbox}{valign=t}
\begin{tabular}{c}
\includegraphics[width=0.185\textwidth]{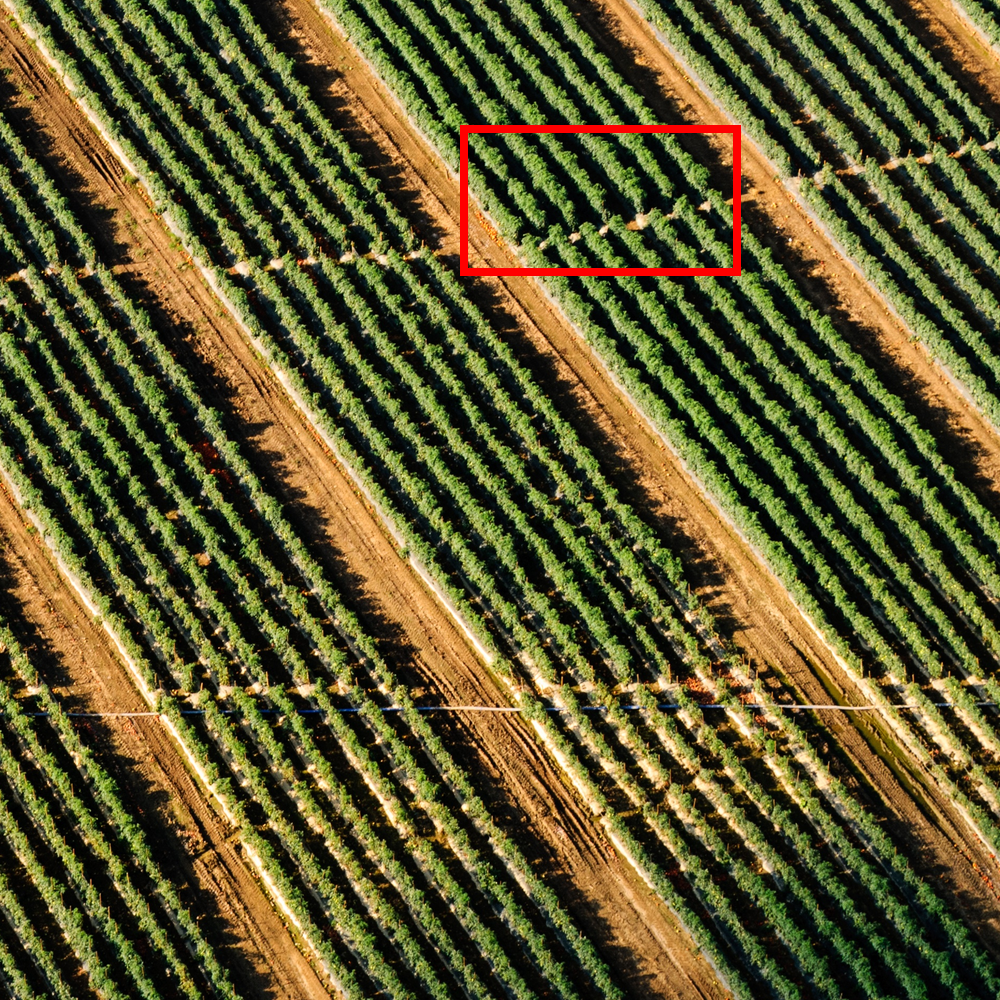}
\\
DIV2K\_val: 0897
\end{tabular}
\end{adjustbox}
\hspace{-0.46cm}
\begin{adjustbox}{valign=t}
\begin{tabular}{cccccc}
\includegraphics[width=0.154\textwidth]{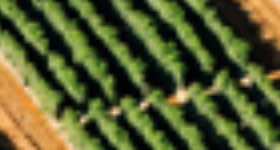} \hspace{-3.5mm} &
\includegraphics[width=0.154\textwidth]{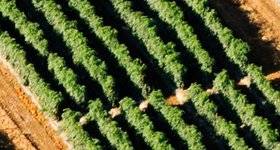} \hspace{-3.5mm} &
\includegraphics[width=0.154\textwidth]{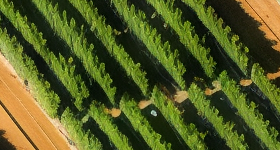} \hspace{-3.5mm} &
\includegraphics[width=0.154\textwidth]{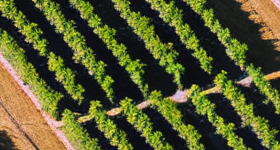} \hspace{-3.5mm} &
\includegraphics[width=0.154\textwidth]{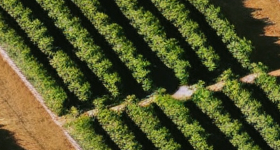} \hspace{-3.5mm} &
\\ 
LR \hspace{-3.5mm} &
HR \hspace{-3.5mm} &
DiffBIR~\cite{diffbir} / 32-bit\hspace{-3.5mm} &
OSEDiff~\cite{OSEDiff} / 32-bit \hspace{-3.5mm} &
PassionSR-FP / 32-bit \hspace{-3.5mm} &
\\
\includegraphics[width=0.154\textwidth]{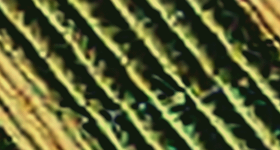} \hspace{-3.5mm} &
\includegraphics[width=0.154\textwidth]{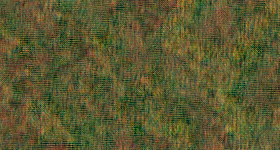} \hspace{-3.5mm} &
\includegraphics[width=0.154\textwidth]{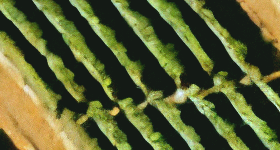} \hspace{-3.5mm} &
\includegraphics[width=0.154\textwidth]{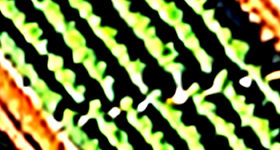} \hspace{-3.5mm} &
\includegraphics[width=0.154\textwidth]{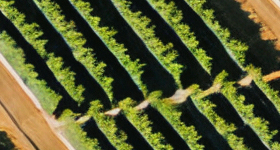} \hspace{-3.5mm} &
\\ 
MaxMin~\cite{choukroun2019lowbitquantizationneuralnetworks} / 8-bit \hspace{-3.5mm} &
LSQ~\cite{esser2019learned}  / 8-bit \hspace{-3.5mm} &
Q-Diffusion~\cite{q-diffusion}  / 8-bit \hspace{-3.5mm} &
EfficientDM~\cite{EfficientDM}  / 8-bit \hspace{-3.5mm} &
PassionSR (ours)  / 8-bit \hspace{-3.5mm}
\\
\end{tabular}
\end{adjustbox}
\end{tabular}
\vspace{-0.5mm}

\begin{tabular}{cccccccc}
\hspace{-0.44cm}
\begin{adjustbox}{valign=t}
\begin{tabular}{c}
\includegraphics[width=0.185\textwidth]{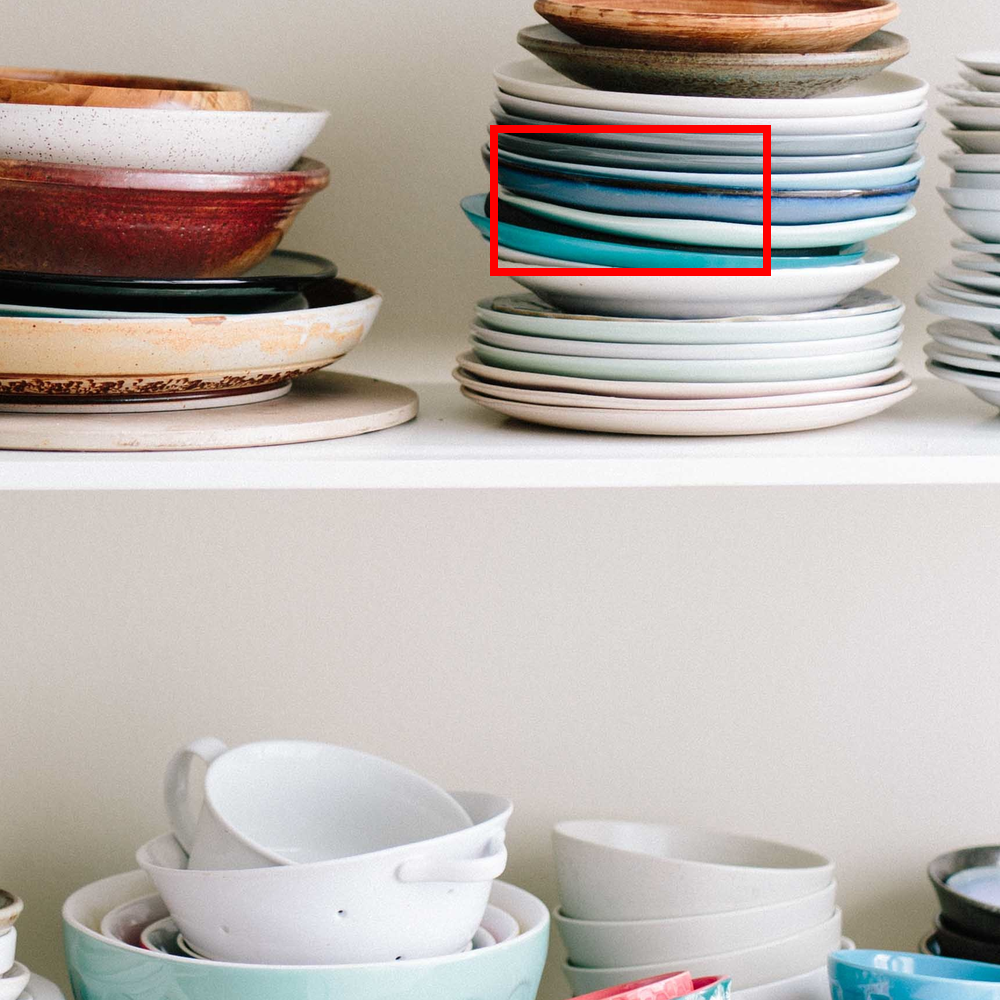}
\\
DIV2K\_val: 0833
\end{tabular}
\end{adjustbox}
\hspace{-0.46cm}
\begin{adjustbox}{valign=t}
\begin{tabular}{cccccc}
\includegraphics[width=0.154\textwidth]{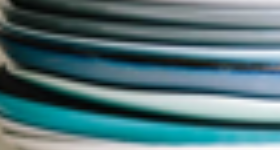} \hspace{-3.5mm} &
\includegraphics[width=0.154\textwidth]{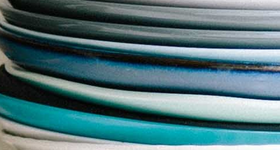} \hspace{-3.5mm} &
\includegraphics[width=0.154\textwidth]{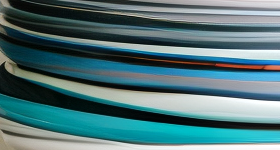} \hspace{-3.5mm} &
\includegraphics[width=0.154\textwidth]{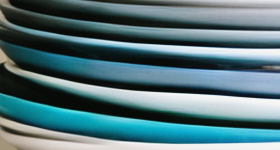} \hspace{-3.5mm} &
\includegraphics[width=0.154\textwidth]{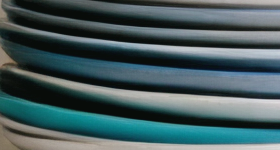} \hspace{-3.5mm} &
\\ 
LR \hspace{-3.5mm} &
HR \hspace{-3.5mm} &
DiffBIR~\cite{diffbir} / 32-bit\hspace{-3.5mm} &
OSEDiff~\cite{OSEDiff} / 32-bit \hspace{-3.5mm} &
PassionSR-FP / 32-bit \hspace{-3.5mm} &
\\
\includegraphics[width=0.154\textwidth]{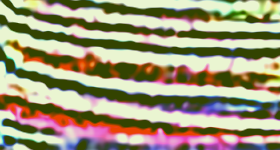} \hspace{-3.5mm} &
\includegraphics[width=0.154\textwidth]{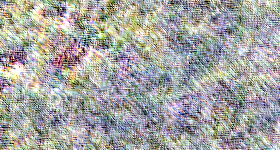} \hspace{-3.5mm} &
\includegraphics[width=0.154\textwidth]{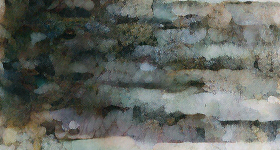} \hspace{-3.5mm} &
\includegraphics[width=0.154\textwidth]{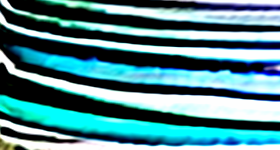} \hspace{-3.5mm} &
\includegraphics[width=0.154\textwidth]{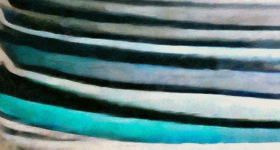} \hspace{-3.5mm} &
\\ 
MaxMin~\cite{choukroun2019lowbitquantizationneuralnetworks} / 6-bit \hspace{-3.5mm} &
LSQ~\cite{esser2019learned}  / 6-bit \hspace{-3.5mm} &
Q-Diffusion~\cite{q-diffusion}  / 6-bit \hspace{-3.5mm} &
EfficientDM~\cite{EfficientDM}  / 6-bit \hspace{-3.5mm} &
PassionSR (ours)  / 6-bit \hspace{-3.5mm}
\\
\end{tabular}
\end{adjustbox}
\end{tabular}
\vspace{-0.5mm}

\begin{tabular}{cccccccc}
\hspace{-0.44cm}
\begin{adjustbox}{valign=t}
\begin{tabular}{c}
\includegraphics[width=0.185\textwidth]{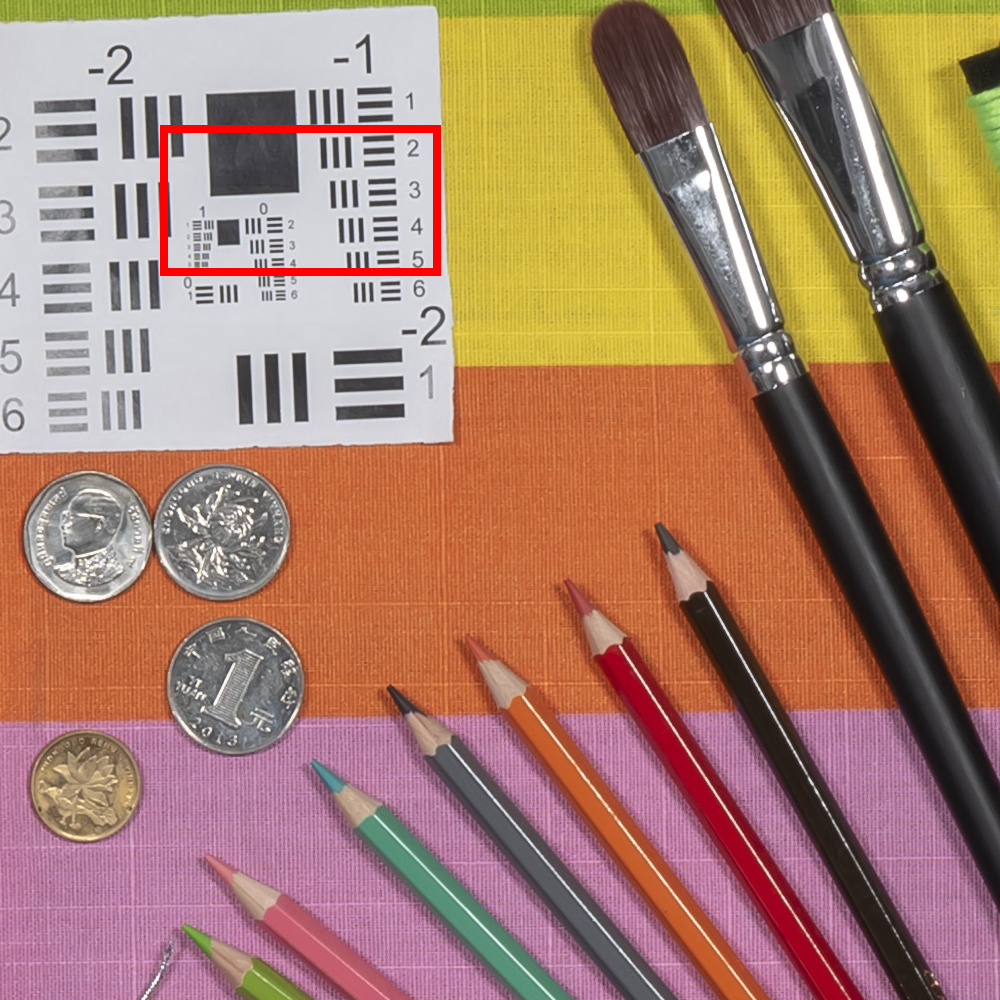}
\\
RealSR: Nikon\_049
\end{tabular}
\end{adjustbox}
\hspace{-0.46cm}
\begin{adjustbox}{valign=t}
\begin{tabular}{cccccc}
\includegraphics[width=0.154\textwidth]{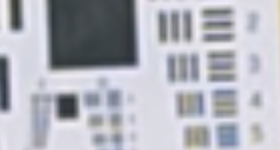} \hspace{-3.5mm} &
\includegraphics[width=0.154\textwidth]{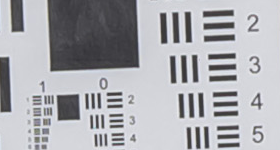} \hspace{-3.5mm} &
\includegraphics[width=0.154\textwidth]{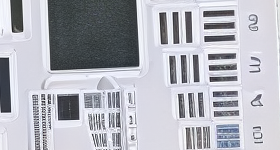} \hspace{-3.5mm} &
\includegraphics[width=0.154\textwidth]{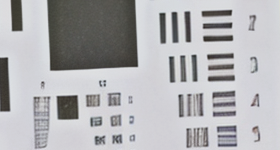} \hspace{-3.5mm} &
\includegraphics[width=0.154\textwidth]{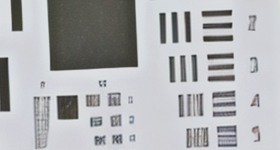} \hspace{-3.5mm} &
\\ 
LR \hspace{-3.5mm} &
HR \hspace{-3.5mm} &
DiffBIR~\cite{diffbir} / 32-bit \hspace{-3.5mm} &
OSEDiff~\cite{OSEDiff} / 32-bit \hspace{-3.5mm} &
PassionSR-FP / 32-bit \hspace{-3.5mm} &
\\
\includegraphics[width=0.154\textwidth]{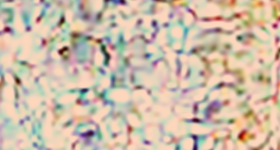} \hspace{-3.5mm} &
\includegraphics[width=0.154\textwidth]{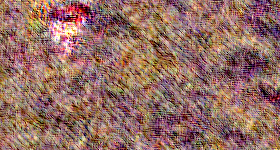} \hspace{-3.5mm} &
\includegraphics[width=0.154\textwidth]{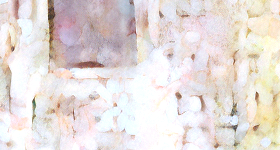} \hspace{-3.5mm} &
\includegraphics[width=0.154\textwidth]{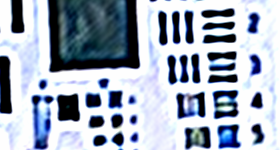} \hspace{-3.5mm} &
\includegraphics[width=0.154\textwidth]{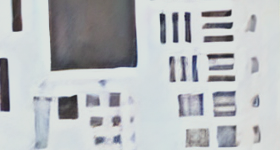} \hspace{-3.5mm} &
\\ 
MaxMin~\cite{choukroun2019lowbitquantizationneuralnetworks} / 6-bit \hspace{-3.5mm} &
LSQ~\cite{esser2019learned}  / 6-bit \hspace{-3.5mm} &
Q-Diffusion~\cite{q-diffusion}  / 6-bit \hspace{-3.5mm} &
EfficientDM~\cite{EfficientDM}  / 6-bit \hspace{-3.5mm} &
PassionSR (ours)  / 6-bit \hspace{-3.5mm}
\\
\end{tabular}
\end{adjustbox}
\end{tabular}
\vspace{-4.mm}
\caption{Visual comparison ($\times 4$) with high-resolution image, full-precision model's output and different quantization methods in some challenging cases at W8A8 and W6A6 \textbf{UNet-VAE quantization}. PassionSR gains significant visual advantages over other methods.}
\vspace{-4.mm}
\label{fig:visual-whole-model}
\end{figure*}

\vspace{-3mm}
\section{Experiments}
\vspace{-1mm}
\subsection{Experiment Setup}
\vspace{-1mm}
\noindent\textbf{Data Construction.} We randomly crop 500 LR and HR pairs, each of size 128$\times$128, from DIV2K\_train~\cite{Agustsson_2017_CVPR_Workshops} to construct the calibration dataset. For the test datasets, we select RealSR~\cite{Ji_2020_CVPR_Workshops}, DRealSR~\cite{wei2020cdc}, and DIV2K\_val~\cite{Agustsson_2017_CVPR_Workshops}.

\begin{table}[t]
\centering
\vspace{1.5mm}
\resizebox{1\columnwidth}{!}{
\begin{tabular}{c | c | c c}
\toprule
\rowcolor{cvprblue!30}
\textbf{Method} & \textbf{Bit} & \textbf{Params / M} ($\downarrow$ Ratio)  & \textbf{Ops / G} ($\downarrow$ Ratio)\\
\midrule
OSEDiff      & W32A32 & 1,303 ($\downarrow$0\%) & 4,523 ($\downarrow$0\%) \\
PassionSR-FT & W32A32 & 949 ($\downarrow$27.13\%) & 4,240 ($\downarrow$6.25\%) \\
\midrule
& W8A8 & 300 ($\downarrow$76.96\%) & 3,732 $\downarrow$17.50\%) \\
\multirow{-2}{*}{PassionSR-U}
& W6A6 & 246 ($\downarrow$81.11\%) & 3,689 ($\downarrow$18.44\%) \\
\midrule
\rowcolor{cvprblue!10}
& W8A8 & 238 ($\downarrow$81.77\%) & 1,060 ($\downarrow$76.56\%) \\
\rowcolor{cvprblue!10}
\multirow{-2}{*}{PassionSR-UV}
& W6A6 & 178 ($\downarrow$86.32\%) & 795 ($\downarrow$82.42\%) \\
\bottomrule
\end{tabular}
}
\vspace{-4mm}
\caption{Compression ratio of different quantization settings. PassionSR-U refers to UNet-only quantization while PassionSR-UV refers to UNet-VAE quantization.}
\label{tab:Compression ratio}
\vspace{-6mm}
\end{table}

\noindent\textbf{Evaluation Metrics.} We employ reference-based evaluation metrics, including PSNR, SSIM~\cite{wang2004image}, LPIPS~\cite{zhang2018unreasonable}, and DISTS~\cite{ding2020image}. We also utilize non-reference metrics, such as NIQE~\cite{zhang2015feature}, MUSIQ~\cite{ke2021musiq}, ManIQA~\cite{yang2022maniqa}, and ClipIQA~\cite{wang2023exploring}. We evaluate all methods with full-size images. 

\noindent\textbf{Implementation Details.} We quantize the weights and activations in the main components (\ie, UNet and VAE) with low bit-widths (\eg, 6 and 8 bits). We denote the quantization configuration $w$-bit weight quantization and $a$-bit activation quantization as $\text{W}w\text{A}a$. For calibration training, we set the learning rate of PassionSR as 1$\times$${10}^{-5}$ and finetune for 4 epochs. It is worth demonstrating that we use the same initialization method as SmoothQuant~\cite{smoothquant}.

\noindent\textbf{Compared Methods.} We select representative quantization methods: MaxMin~\cite{jacob2017quantizationtrainingneuralnetworks}, LSQ~\cite{esser2019learned}, Q-Diffusion~\cite{q-diffusion}, and EfficientDM~\cite{EfficientDM}. We adopt these methods to quantize our full-precision PassionSR-FP based on their released code.

\begin{table*}[t]
\centering
\scriptsize
\resizebox{\textwidth}{!}{
\begin{tabular}{c | c c | c c c c c c c c}
\toprule
\rowcolor{cvprblue!30}
&  \multicolumn{2}{c|}{Efficiency} & \multicolumn{8}{c|}{RealSR} \\
\rowcolor{cvprblue!30}
\multirow{-2}{*}{Methods} & Time (h) & GPU (GB) & PSNR$\uparrow$  & SSIM$\uparrow$ & LPIPS$\downarrow$ & DISTS$\downarrow$ & NIQE$\downarrow$ & MUSIQ$\uparrow$ & MANIQA$\uparrow$ & CLIP-IQA$\uparrow$ \\
\midrule
MaxMin & 0.00 & 0 &15.55 &0.2417	&0.8018	&0.4449	&9.263	&42.15	&0.2791	&0.4174 \\
LBQ  & 2.66 & 40 & 23.15 & 0.6621 & 0.5022 & 0.3115 & 7.234 & 47.75 & 0.3071 & 0.4787 \\
LBQ+LET & 3.87 & 40 & \textcolor{red}{25.40} & \textcolor{red}{0.7529} & 0.3798 & 0.2584 & 6.604 & 44.26 & 0.2414 & 0.3224 \\
\rowcolor{cvprblue!10}
LBQ+LET+DQC & 1.07 & 28 & 24.41 & 0.7374 & \textcolor{red}{0.3427} & \textcolor{red}{0.2419} & \textcolor{red}{5.449} & \textcolor{red}{55.08} & \textcolor{red}{0.3083} & \textcolor{red}{0.4849} \\
\bottomrule
\end{tabular}
}
\vspace{-3.mm}
\caption{Ablation study on our proposed components: LBQ, LET, and DQC. Our ablation experiments are in the setting of W6A6 UNet-VAE quantization. We test each ablation method on RealSR and record their calibration time and GPU costs.}
\vspace{-6.mm}
\label{tab:ablation experiments}
\end{table*}

\begin{figure*}[t]
    \centering
    \begin{minipage}{0.32\textwidth}
        \centering
        \includegraphics[width=\linewidth]{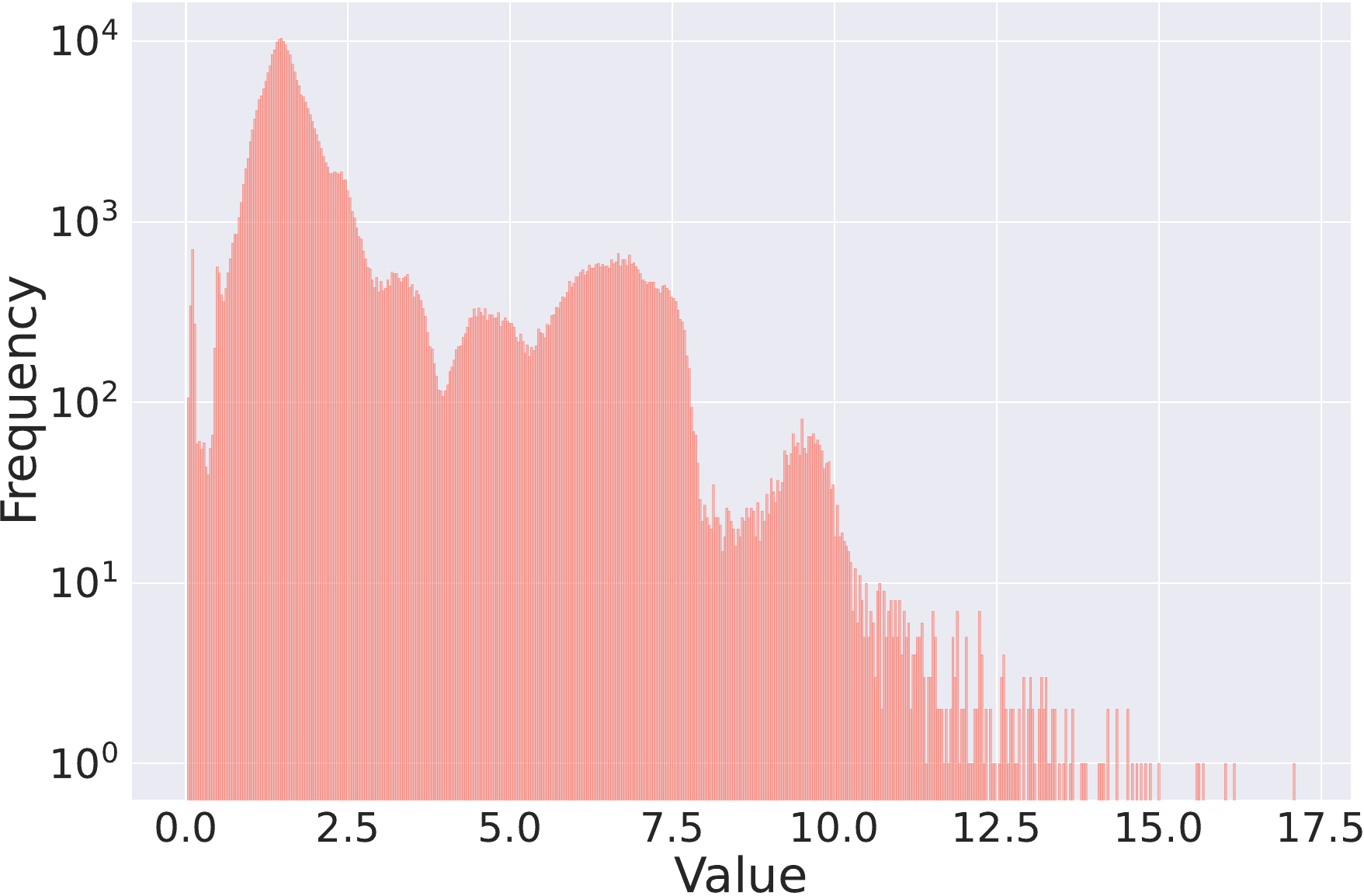}
        \subcaption{Distribution of scale factor}
        \label{fig:Distribution of scale factor}
    \end{minipage}
    \hfill
    \begin{minipage}{0.32\textwidth}
        \centering
        \includegraphics[width=\linewidth]{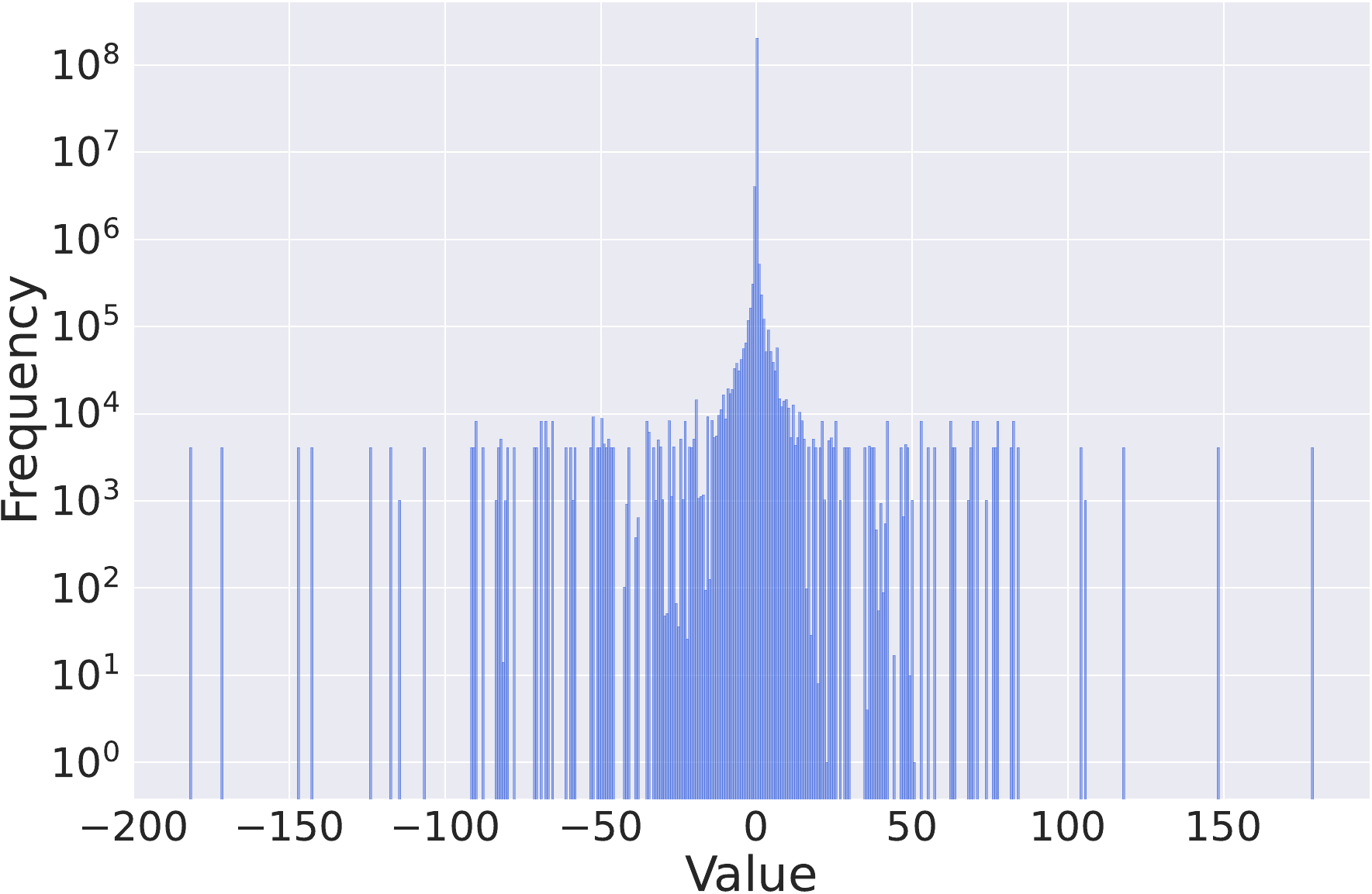}
        \subcaption{Distribution of Original Activation}
        \label{fig:Distribution of Original Activation}
    \end{minipage}
    \hfill
    \begin{minipage}{0.32\textwidth}
        \centering
        \includegraphics[width=\linewidth]{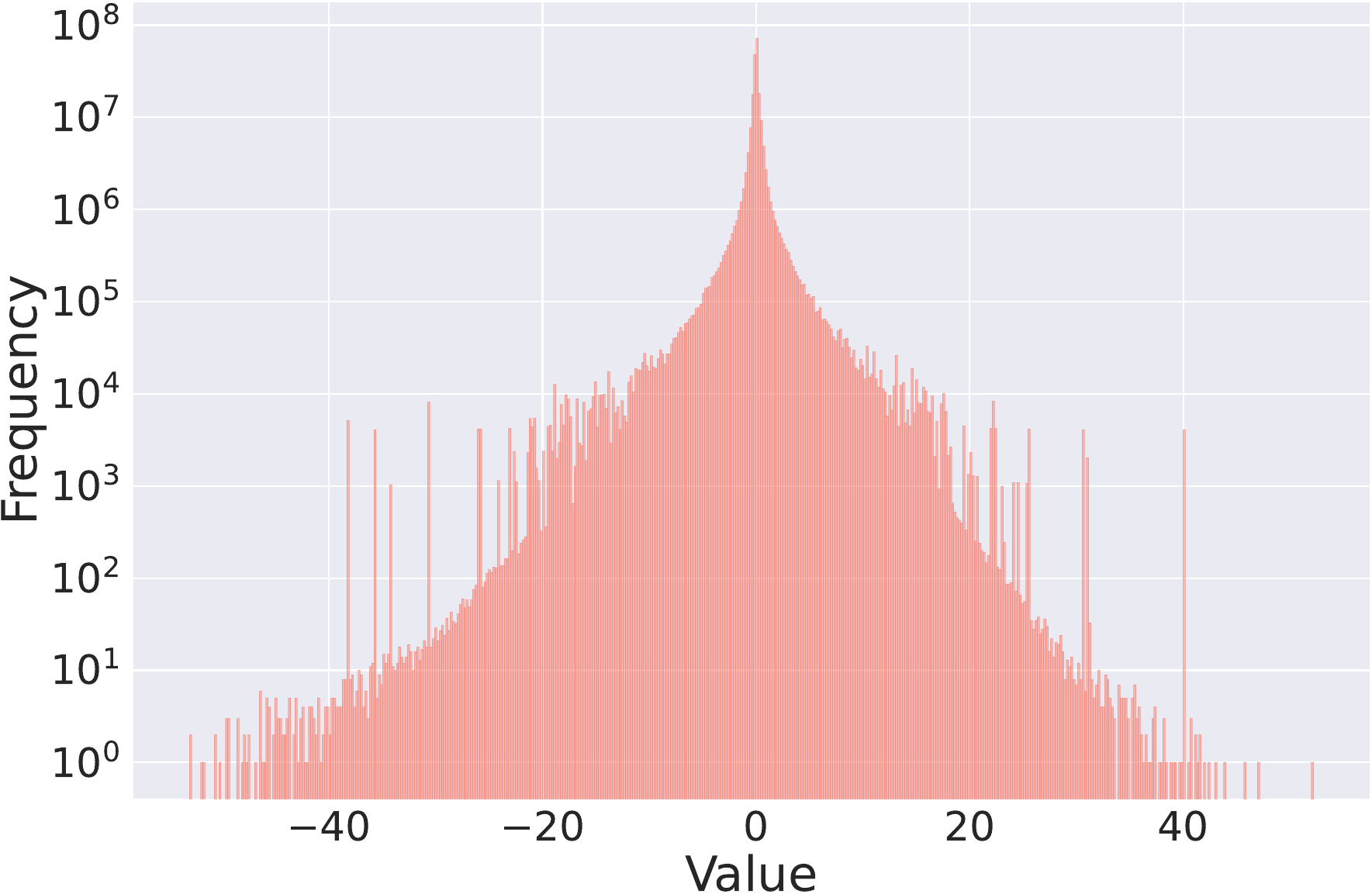}
        \subcaption{Distribution of Smooth Activation}
        \label{fig:Distribution of Smooth Activation}
\end{minipage}
\vspace{-3.5mm}
\caption{Distribution of scale factor and activation before and after smooth in the whole model.}
\label{fig:Distribution-plot}
\vspace{-5mm}
\end{figure*}

\vspace{-1mm}
\subsection{Main Results}
\vspace{-1mm}
\textbf{Streamline Experiment.}
We replace the text embedding branch with a constant empty prompt embedding, preprocessed by the ClipEncoder using an empty string. Table~\ref{tab:Whole model quantization experiment results.} shows a comparison between the original model OSEDiff and the streamline model PassionSR-FP. The results indicate minimal or even negligible performance drop. Additionally, as shown in Tab.~\ref{tab:Compression ratio}, the total model parameters are reduced by 27.13\%, and operations are reduced by 6.25\%.

\noindent\textbf{Quantitative Results.} In the UNet-VAE quantization experiment, Tab.~\ref{tab:Whole model quantization experiment results.} shows that PassionSR significantly outperforms previous methods at the setting of W8A8 and W6A6. On each dataset, 8-bit PassionSR achieves comparable performance to the full-precision OSEDiff, or even better scores in some cases. Besides PassionSR, other quantization methods have an obvious decrease in quantitative results. For 6-bit quantization, while contrast methods like LSQ and Q-Diffusion exhibit low structural metrics (\eg, PSNR and SSIM), they obtain high scores in non-reference IQA metrics. Our PassionSR achieves the highest reference IQA values and relatively lower non-reference IQA values than others. However, we find an interesting observation where images with substantial quantization noise also obtain relatively high non-reference IQA scores. It can explain why some quantized outputs by other methods have worse visual quality despite higher non-reference IQA scores. We provide more results and analyses in Fig.~\ref{fig:visual-whole-model}.

\noindent\textbf{Visual Comparison.} We provide a visual comparison ($\times$4) for UNet-VAE quantization in Fig.~\ref{fig:visual-whole-model}. Several challenging cases are selected for clearer visual contrast. Compared to previous methods, PassionSR generates clearer results and better textures, with a minimal gap from the full-precision model. Notably, PassionSR even surpasses the full-precision PassionSR-FP in certain cases. 

\noindent\textbf{Compression Ratio.} To clearly present our compression and acceleration outcomes, we calculate the model’s total size (Params / G) and the number of operations (Ops / G). The calculation  follows the methods used in previous quantization studies~\cite{quantsr}. The results are presented in Tab.~\ref{tab:Compression ratio}, showing the compression and acceleration ratios for each setting. In the 8-bit setting, PassionSR-UV achieves an 81.77\% compression ratio and a 76.56\% acceleration ratio. Furthermore, with a 6-bit setting, we achieve a compression ratio of 86.32\% and an acceleration ratio of 82.42\%.

\vspace{-2mm}
\subsection{Ablation Study}
\vspace{-2mm}
\textbf{Learnable Equivalent Transformation (LET).} In order to evaluate the effects of LET, we adopt the LBQ-only quantizer without LET to make comparison with LBQ and LET combined quantizer. The experiment results is presented in Tab.~\ref{tab:ablation experiments}. Considering that the output images have lots of noise, the reference IQA metrics are more reasonable to measure the visual quality. The introduction of LET improves PSNR by over 2 dB and SSIM by about 0.1, with an additional hour of training, which indicates that LET brings huge performance increase. 

\noindent\textbf{Distributed Quantization Calibration (DQC).} To research into the special calibration strategy DQC, we use collaborative calibration with LBQ and LET combined quantizer compared with DQC. The detailed experiments results are listed in Tab.~\ref{tab:ablation experiments}. With slight performance enhancement, the integration of DQC mainly leads to faster converge and lower GPU memory cost,  reducing the calibration time by about 2 hours and the GPU memory by over 10 G. It is because that the amount of parameters requiring gradients is smaller when applied DQC and the computational costs reduce during backward propagation. With distributed calibration, the training of scale factor in LET becomes stable, making it easier to find out the best parameters in LET.

\subsection{Distribution Visualization}
\vspace{-1mm}
By applying the equivalent transformation to UNet and VAE, we can adjust the values of scale factors to better control the distributions of weights and activations, making them easier to quantize. LET playes an important role in the quantization and we visualize the distribution of relative variables to observe LET's detailed effects in Fig.~\ref{fig:Distribution-plot}. 

\noindent\textbf{Distribution of Scale Factor.} The dispersed distribution of scale factors in Fig.~\ref{fig:Distribution of scale factor} implies that the scale factors in different channel or tensor take different values to address varying activation distributions. We obtain the best scale factor for different layer with different activation distributions through calibration process. Most of scale factor are larger than 1, which means that LET mainly alleviates the difficulties in activation quantization.

\noindent\textbf{Distribution of Activation.} Figure~\ref{fig:Distribution of Original Activation} shows that the distribution of original activations before LET is truly dispersed. A large amount of outliers have severe impacts on quantization. Under the help of LET, the distribution of activations is more centered and more friendly for quantization as shown in Fig.~\ref{fig:Distribution of Smooth Activation}. Moreover, the number of outliers largely decreases. This change in activation distribution indicates LET's strong ability to manipulate the distribution and great effects on minimizing the quantization errors.

\vspace{-2mm}
\section{Conclusion}
\vspace{-2mm}

In this paper, we propose PassionSR, a novel post-training quantization method for one-step diffusion-based image super-resolution. By simplifying the model architecture to two core components, Variational Autoencoder (VAE) and UNet, we obtain a UNet-VAE structure with little performance loss. Our approach incorporates a Learnable Boundary Quantizer (LBQ) and Learnable Equivalent Transformation (LET) to manipulate activation distributions. And Distributed Quantization Calibration (DQC) strategy enhances training stability and accelerates convergence. Experiments show that PassionSR delivers perceptual performance comparable to full-precision models at 8-bit and 6-bit. It gains significant advantages over recent leading diffusion quantization methods. This work paves the way for future one-step diffusion-based SR model quantization and practical deployment of advanced SR model applications. 

\newpage
\bibliographystyle{ieeenat_fullname}
\bibliography{main}

\end{document}